\def\year{2021}\relax
\documentclass[letterpaper]{article} % DO NOT CHANGE THIS
\usepackage{aaai21}  % DO NOT CHANGE THIS
\usepackage{times}  % DO NOT CHANGE THIS
\usepackage{helvet} % DO NOT CHANGE THIS
\usepackage{courier}  % DO NOT CHANGE THIS
\usepackage[hyphens]{url}  % DO NOT CHANGE THIS
\usepackage{graphicx} % DO NOT CHANGE THIS
\usepackage{natbib}  % DO NOT CHANGE THIS AND DO NOT ADD ANY OPTIONS TO IT
\usepackage{caption} % DO NOT CHANGE THIS AND DO NOT ADD ANY OPTIONS TO IT
\usepackage{amsmath}
\usepackage{amssymb}
\usepackage{graphicx}
\usepackage{caption}
\usepackage{color}
\usepackage{multirow}
\usepackage{bm}
\usepackage{adjustbox}
\usepackage{bbold}
\usepackage{wrapfig}
\usepackage{adjustbox}
\usepackage[linesnumbered,ruled]{algorithm2e}
\usepackage{array}
\usepackage{tabularx}
\newcolumntype{L}[1]{>{\raggedright\arraybackslash}p{#1}}
\newcolumntype{R}[1]{>{\raggedleft\arraybackslash}p{#1}}
\newcommand{\RN}[1]{%
  \textup{\uppercase\expandafter{\romannumeral#1}}%
}
\usepackage{floatrow}
\usepackage{graphicx}
\usepackage{caption}
\usepackage[labelformat=simple]{subcaption}
\usepackage{color}
\usepackage{multirow}

\usepackage[belowskip=-4pt,aboveskip=0pt]{caption}
\usepackage{tabularx}
\newcolumntype{C}{>{\Centering\arraybackslash}X} % centered "X" column

\title{Type-augmented Relation Prediction in Knowledge Graphs}
\author{Zijun Cui\textsuperscript{\rm 1}, Pavan Kapanipathi\textsuperscript{\rm 2}, Kartik Talamadupula\textsuperscript{\rm 2}, Tian Gao\textsuperscript{\rm 2}, Qiang Ji\textsuperscript{\rm 1} \\ %Note that the comma should be placed BEFORE the superscript for optimum readability % email address must be in roman text type, not monospace or sans serif
}
\affiliations{
\textsuperscript{\rm 1}Rensselaer Polytechnic Institute, \textsuperscript{\rm 2} IBM Research\\
\textsuperscript{\rm 1}\{cuiz3, jiq\}@rpi.edu, 
\textsuperscript{\rm 2}\{kapanipa, krtalamad, tgao\}@us.ibm.com}

\begin{document}

\maketitle

%%%%%%%%% ABSTRACT
\begin{abstract}
Knowledge graphs (KGs) are of great importance to many real world applications, but they generally suffer from incomplete information in the form of missing relations between entities. Knowledge graph completion (also known as relation prediction) is the task of inferring missing facts given existing ones. Most of the existing work is proposed by maximizing the likelihood of observed instance-level triples. Not much attention, however, is paid to the ontological information, such as type information of entities and relations. In this work, we propose a type-augmented relation prediction (TaRP) method, where we apply both the type information and instance-level information for relation prediction. In particular, type information and instance-level information are encoded as prior probabilities and likelihoods of relations respectively, and are combined by following Bayes' rule. Our proposed TaRP method achieves significantly better performance than state-of-the-art methods on four benchmark datasets: FB15K, FB15K-237, YAGO26K-906, and DB111K-174. In addition, we show that TaRP achieves significantly improved data efficiency. More importantly, the type information extracted from a specific dataset can generalize well to other datasets through the proposed TaRP model.
\end{abstract}

%structured representation of real world facts, and 

%State-of-the-art relation prediction methods are mainly knowledge graph embedding-based methods, and 

%Such type information is available in many knowledge graphs, e.g. FreeBase, YAGO, and DBpedia. 
%we first encode the type information as prior probabilities. %and derive the prior probability distribution of relations conditioned on a pair of entities based on the semantic similarity. 
%The existing KG-embedding based models are employed to model the likelihoods of relations based on instance-level information. An integration procedure is then proposed to obtain the posterior probabilities, where prior probabilities are combined with the likelihoods by following the Bayes rule. 

%%%%%%%%% BODY TEXT
\section{Introduction}
\label{sec:intro}
\begin{figure}[ht]
    \centering
    \includegraphics[width=3in, height = 2in]{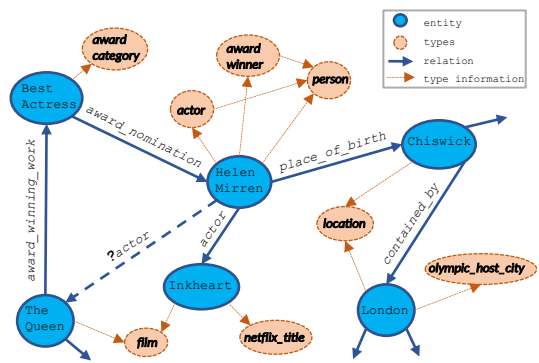}
    \caption{An example in a knowledge graph (KG).}
    \label{fig:KGexample}
\end{figure}

%In the era of knowledge engineering, a large number of knowledge graphs (KGs) have been developed, such as YAGO~\cite{rebele2016yago}, FreeBase~\cite{bollacker2008freebase}, and Google Knowledge Graph\footnote{https://developers.google.com/knowledge-graph}. 
Knowledge graphs (KGs) have gained significant popularity due to successful applications to many different AI tasks such as question answering~\cite{huang2019knowledge}, recommendation~\cite{wang2019kgat}, dialogue generation~\cite{xu2020dynamic}, and natural language inference~\cite{Wang2019,kapanipathi2019infusing}.
% Knowledge graphs (KGs) store efficiently structured knowledge of real-world named entities, facts, or concepts. 
%Formally, let $\mathcal{E}$ denote the entity set and $\mathcal{R}$ denote the relation set. A KG $\mathcal{G}$ usually is a collection of a large set of triplets $(e_h,r,e_t)$, indicating a head entity $e_h$ connected to a tail entity $e_t$ through a relation $r$, where $\{e_h, e_r\}\in \mathcal{E}$, $r \in \mathcal{R}$. 
%For example, a triple (\texttt{Helen\_Mirren}, \texttt{place\_of\_birth}, \texttt{Chiswick}) is represented as two entities: \texttt{Helen\_Mirren} and \texttt{Chiswick} along with a relation \texttt{place\_of\_birth} indicating Helen is born in Chiswick. 
However, KGs are generally incomplete and suffer from missing relations between entities~\cite{socher2013reasoning, west2014knowledge}. The task of knowledge graph completion or relation prediction is aimed at tackling this issue, i.e., inferring missing facts given existing ones. For example, in Figure~\ref{fig:KGexample}, given two entities, e.g., $\texttt{Helen\_Mirren}$ and $\texttt{The\_Queen}$, the relation prediction task predicts if those entities are connected by any of the existing relations in the KG, e.g., $\texttt{actor}$.  %\textcolor{blue}{talk about why don't perform entity prediction /justifications form other relation prediction works}

Relation prediction methodologies are mostly based on KG embeddings, and can primarily be categorized based on the two kinds of information they use from KGs: (i) Instance-level information, i.e., existing triples connecting entities through relations, such as \texttt{Helen\_Mirren $\rightarrow$ place\_of\_birth $\rightarrow$ Chiswick}; and (ii) Ontological information, i.e., meta information about entities and relations, such as the type information of entities %from the domain and range of a relation is used: for 
e.g., \texttt{Helen\_Mirren} is of types \textbf{\textit{\{actor, award\_winner, person\}}}. %$\rightarrow$ type $\rightarrow$} \textbf{\textit{actor}} and \textbf{\textit{actor}}\texttt{$\rightarrow$ is\_a $\rightarrow$} \textbf{\textit{person}}. 
The majority of existing methods use merely instance-level information for learning the embeddings~\cite{sun2019rotate,zhang2019quaternion}, while a few other models use both instance-level information and ontological information~\cite{hao2019universal,garg2019quantum,xie2016representationD}.
%While most methods use instance-level data~\cite{} for learning the embeddings, 
Ontological information such as type information can intuitively help relation prediction, as most relations may connect two distinct types of entities as domain and range. For example, the relation \texttt{place\_of\_birth} always connects entities of type \textbf{\textit{person}} to entities of type \textbf{\textit{location}}. Integrating such type information into instance-level training triples can benefit the relation prediction task, in particular when there is a lack of sufficient training data for learning embeddings. 

A few existing embedding based models with such type information integrated have shown success~\cite{guo2015semantically,xie2016representationT,ma2017transt,garg2019quantum,hao2019universal}. However, these models integrate the ontological information through the model training procedure for better learning the embeddings, and are hence prone to the following drawbacks: (i) the type information is not explicitly differentiated from the instance-level information, and a single set of model parameters are learned by considering two kinds of information jointly; %the approaches learn the same parameters for both kinds of data~\cite{}; 
(ii) the type information is tightly encoded into the objective function, making the integration highly reliant on the training procedure and hence less flexible in augmenting new embedding techniques. We refer to such integration procedures as {\em feature-level integration}. Instead, we proposed an effective {\em decision-level integration}: given the type information and instance-level information encoded as prior probabilities and likelihoods respectively, the proposed decision-level integration combines the two kinds of information by following Bayes' rule. %where the type information and the instance-level information are encoded as prior probabilities and likelihoods respectively and are combined by following the Bayes rule. %. KG embedding based methods encode KG structures into low-dimensional embedding spaces where the symbolic representations of entities and relations are represented as continuous embedding vectors. 
In this paper, we propose a simple but effective framework to augment existing embedding based models with type information. %% What the approach does. 
The contributions of our work are as follows:
\begin{itemize}
%% Novel, simple and effective framework for XYZ. We show that it improves performance
%% Less data experiments
    %\item We propose to encode the type information %from a relation's domain and range as prior probabilities. In particular, we introduce hierarchy-based type weights to encode not only the semantic types of entities and relations, but also the underlying hierarchy structures among types.
    
    \item The proposed decision-level integration framework is independent of the embedding based model, and can be flexibly applied for augmenting different embedding based models without additional training. 
    
    \item The proposed type-augmented relation prediction (TaRP) method achieves better relation prediction performance than state-of-the-art models on three benchmark datasets. Furthermore, we show that by incorporating the type information, TaRP has less dependency on training data, and thus is more data efficient.
    
    \item We empirically demonstrate that the type information extracted from a specific dataset can generalize well to other, different datasets through the proposed TaRP model.
    
    %achieves better relation prediction performance than state-of-the-art models on three benchmark datasets. In addition, we show that by incorporating the type information, TaRP has less dependency on training data, and thus is more data efficient. More importantly, we show that the type information extracted from a certain dataset can generalize well to other datasets. %perform cross-dataset evaluation demonstrating that our prior model can generalize well to different datasets. To the best of our knowledge, we are the first one evaluating the generalization ability of the prior knowledge on different datasets.
\end{itemize}

\section{Related Work}
\label{sec:related_work}

KG embedding based methods have been widely explored for the KG completion task. The general methodology of the embedding based method is to define a score function for triples within a continuous embedding space. The score function usually takes the form $f_r(\textbf{e}_h, \textbf{e}_t)$, where $\textbf{e}_h, \textbf{e}_t$ are head and tail entity embeddings. The score function measures the salience of a candidate triple $(e_h, r, e_t)$, and embeddings of entities and relations are learned by optimizing the score function. TransE~\cite{bordes2013translating} represents entities and relations in $d$-dimensional vector space, i.e., $\textbf{e}_h, \textbf{e}_r, \textbf{r} \in R^d$ and learns the embeddings by assuming the translation principle $\textbf{e}_h + \textbf{r} \approx \textbf{e}_t$ with the proposed score function $f_r(\textbf{e}_h, \textbf{e}_t) = -||\textbf{e}_h + \textbf{r}-\textbf{e}_t||$. %DistMult~\cite{yang2014embedding} considered semantic similarity and associated related entites using Hadamard product of embeddings. ComplEx~\cite{trouillon2016complex} extended DistMult in a complex space and produced comparable performance. 
Along the lines of the translation-based methods (first introduced by TransE), many more advanced methods have been proposed, such as TransH~\cite{wang2014knowledge} and TransD~\cite{ji2015knowledge}.
More recently, RotatE~\cite{sun2019rotate} and QuatE~\cite{zhang2019quaternion} %And QuatE~\cite{zhang2018knowledge} is the current state-of-the-art embedding-based method for knowledge graph completion. 
have been proposed by representing the entities and relations using complex vectors.
In addition, neural networks have also been introduced to learn robust embedding based models, such as ConvE~\cite{dettmers2018convolutional} and HypER~\cite{balavzevic2019hypernetwork}. These methods learn embeddings  based on instance-level information observed from existing triples, without considering the rich ontological information that exists. 

There are existing embedding based methods that explore the usage of the ontological information. DistMult~\cite{yang2014embedding} considered semantic similarity and associated related entities using Hadamard product of embeddings.  \citet{guo2015semantically} proposed semantically smooth embedding, where the type information is encoded as smoothness constraints. %Entities of the same type are assumed to be close to each other in the embedding space, and the type information is then encoded as smoothness constraints. 
\citet{ma2017transt} measured the type-based semantic similarity between entities and relations, and that semantic similarity served as the prior probability. Besides semantic types of entities, underlying hierarchy structures among types are also considered. \citet{xie2016representationT} proposed type-specific entity projections by applying hierarchical type information, and devised type-embodied knowledge representation learning (TKRL). \citet{hao2019universal} introduced instance-view KG and ontology-view KG, and the hierarchy structures among types are explicitly represented within the ontology-view KG. Universal embeddings are then learned by considering the two types of KGs jointly. \citet{zhang2020learning} proposed hierarchy-aware KG embedding for link prediction. In addition, hierarchical type information is extracted as logical propositions for the quantum embeddings~\cite{garg2019quantum}, %proposed quantum embedding, by considering all logical propositions from A-box and T-box.
and symbolic KGs are represented with embedding vectors in a logic structure preserving manner. \citet{jain2018type} considered type information for entity prediction without explicit supervision. %. ComplEx~\cite{trouillon2016complex} extended DistMult in a complex space and produced comparable performance. Entity types are also applied as constraints on head and tail positions for different relations~\cite{krompass2015type}. In addition, relations are categorized into different groups based on the observation from the data. Lin et al~\cite{lin2016knowledge} proposed to categorize relation types into attributes and relations. Zhang et al~\cite{zhang2018knowledge} proposed a three-layer hierarchical relation structure containing relation clusters, relations and sub-relations. 
Besides the type information, other ontological information is also explored. DKRL~\cite{xie2016representationD} applies entity descriptions, which are represented as vectors. SSP~\cite{xiao2017ssp} uses the topic distribution of entity descriptions to construct semantic hyperplanes. All of these models integrate the ontological information into the instance-level information at the feature-level in order to learn the embeddings better.

Besides embedding based methods, path based methods have also been proposed for thhe KG completion task~\cite{lao2011random,das2016chains,chen2018variational,zhang2020mcmh}. %, e.g.,~\cite{lao2011random,das2016chains,chen2018variational}. Path based methods infer missing facts based on multi-hop paths going from the head entity to tail entity in KGs, and hence offer logical insights into the underlying KG. Type information is also explored in recent path based methods. 
\citet{lei2019path} utilizes type semantics from the relation to obtain attention that constrains the semantics of the entity. Path based methods suffer from high computational cost because of the path finding procedure; in this paper, we focus on embedding based approaches.

\section{Proposed Method}
\label{sec:proposed_method}

To leverage the type information of entities and relations in improving the relation prediction performance, we propose a type-augmented relation prediction (TaRP) approach.  %Our approach includes three sub-models: a prior model, a likelihood model, and an  integration model. 
We firstly propose a prior model where we encode the type information as prior probabilities. The existing embedding based models are employed to model the data likelihood of relations based on existing instance-level triples. %that learn entity and relation embeddings based on existing triples. Likelihoods of relations are then calculated based on the learned embeddings. 
In the end, we integrate the type information into the embedding based models at the decision level. %proposed to fuse the type information and instance-level information at the decision level. %The integration is performed by following the Bayes rule. 
%The overview of our proposed TaRP method is shown in Figure~\ref{fig:flowchart}. 
%\begin{figure}[h]
%    \centering
%    \includegraphics[width=3.3in, height = 3.8in]{latex/figures/KGC-flowchart.pdf}
%    \caption{The framework of the Type-augmented Relation Prediction method}
%    \label{fig:flowchart}
%\end{figure}

\subsection{Type Information Encoding} We propose a prior model where we encode the type information as prior probabilities. We denote a knowledge graph as $\mathcal{G}=\{\mathcal{E}, \mathcal{R}, \mathcal{T}\}$, where $\mathcal{E}$, $\mathcal{R}$ and $\mathcal{T}$ indicates the entity set, the relation set, and the type set respectively.
We firstly collect the type set regarding to each entity $e\in \mathcal{E}$ and each relation $r\in \mathcal{R}$ separately. %we collect both its head type set $T_{r,head}$ and tail type set $T_{r, tail}$. 
For each triple $(e_h, r, e_t)$, we then define the type-based prior probability of relation $r$ condition on the entity pair $(e_h, e_t)$ by measuring the semantic similarity. In particular, the semantic similarity is calculated based on the correlation between type sets of entities and relations. 

\paragraph{Type collection for entities and relations} Types of entities usually have underlying hierarchy structures, such as the structure among types \textbf{\textit{\{actor, award\_winner, person\}}} in Figure~\ref{fig:KGexample}. To take such a hierarchy into the consideration, we propose to use the hierarchy-based type weights. Given entity $e$, its type set is denoted as $T_e$. A hierarchy structure among a subset of types is denoted as $H = /t_1/t_2/..../t_k/.../t_K$, where $t_k\in T_e$, $K$ is the total number of hierarchy levels, $t_K$ is the most specific semantic type, and $t_1$ is the most general semantic type. Instead of treating the types in the hierarchy $H$ equally, we weight the types based on their locations in the hierarchy. The weight of type $t_k$ regarding to hierarchy $H$ of entity $e$ is defined as,
\begin{equation}
    w_e^H(t_k) = \frac{\exp(k-1)}{\sum_{j=0}^{K-1}\exp(j)}
\end{equation}
Multiple hierarchical structures can exist for one entity with each hierarchy including a subset of types. For example, in Figure~\ref{fig:KGexample}, entity $e=\texttt{Hellen\_Mirren}$ has three hierarchical structures among its types: $H_1 = \textit{\textbf{/person/actor}}$, $H_2 = \textit{\textbf{/person/award\_winner}}$, and $H_3=\textit{\textbf{/person}}$. Type \textbf{\textit{person}} is included in all three hierarchies. Following Eq.1, we have $w_{e}^{H_i}(\textbf{\textit{person}}) = 0.27$,$i=\{1,2\}$ and $w_{e}^{H_3}(\textbf{\textit{person}}) = 1$.

For each type $t\in T_e$, we calculate its hierarchy-based weight as $w_e(t)=\min(w_e^{H_1}(t), w_e^{H_2}(t),...,w_e^{H_N}(t))$, where $N$ is the total number of hierarchies containing type $t$. In the example above, we thus have $w_{e}(\textbf{\textit{person}}) = 0.27$, $w_{e}(\textbf{\textit{actor}}) = 0.73$, and $w_{e}(\textbf{\textit{award\_winner}}) = 0.73$.  %In the end, for each type $t\T_e$, we obtain its hierarchy-based type weight $w_e(t)$.

For the type sets of relation, we consider both the head type set $T_{r,head}$ and tail type set $T_{r,tail}$ defined as
\begin{equation}
\begin{split}
    T_{r,head} &= \cup_{e\in \text{Head}(r)}T_e \\
    T_{r,tail} &= \cup_{e\in \text{Tail}(r)}T_e
\end{split}
\end{equation}
where $\text{Head}(r)=\{e_h|(e_h,r,e_t)\in \mathcal{G},  \forall e_t\in \mathcal{E}\}$ indicating the set of head entities of relation $r$, and $\text{Tail}(r)=\{e_t|(e_h,r,e_t)\in \mathcal{G},  \forall e_h\in \mathcal{E}\}$ indicating the set of tail entities of relation $r$. Correspondingly, we calculate the type weights as 
\begin{equation}
\begin{split}
    w_{r,head}(t) &= \sum_{e\in \text{Head}(r)}w_e(t), \quad \text{for} \ t\in T_{r,head}\\
    w_{r,tail}(t) &= \sum_{e\in \text{Tail}(r)}w_e(t), \quad \text{for} \ t\in T_{r,tail} 
\end{split}
\end{equation}
%Given the hierarchy-based type weights, we not only encode the types of entities and relations, but also the underlying hierarchy structures among types. 

\paragraph{Type-based prior probability} Given a triple $(e_h, r, e_t)\in \mathcal{G}$, we measure the semantic similarity between entities and relations based on the correlation between their type sets. The similarity score $s(\cdot, \cdot)$ is calculated as
\begin{equation}
    \begin{split}
        s(e_h, r) &= \frac{\sum_{t\in T_{r,head}\cap T_{e_h}}w_{r,head}(t)}{\sum_{t\in T_{r,head}}w_{r,head}(t)}\\
        s(e_t, r) &= \frac{\sum_{t\in T_{r,tail}\cap T_{e_t}}w_{r,tail}(t)}{\sum_{t\in T_{r,tail}}w_{r,tail}(t)}
    \end{split}
\end{equation}
where $T_{r,head}\cap T_{e_h} = \{t|t\in T_{r,head} \ \text{and} \ t\in T_{e_h}\}$ and $T_{r,tail}\cap T_{e_t} = \{t|t\in T_{r,tail} \ \text{and} \ t\in T_{e_t}\}$, and $0\leq s(\cdot, \cdot)\leq 1$.
The prior probability $p(r|\mathcal{T}(e_h, e_r, \mathcal{R}))$ is then defined based on similarity scores as, 
\begin{equation}
    p(r|\mathcal{T}(e_h, e_t, \mathcal{R})) \triangleq \frac{s(e_h, r)s(e_t, r)}{\sum_{r'\in \mathcal{R}}s(e_h, r')s(e_t,r')}
\end{equation}
where $\mathcal{T}(e_h, e_t, \mathcal{R})$ denotes the type information related to the entity pair $(e_h, e_t)$ and the relation set $\mathcal{R}$. To obtain a valid prior probability, for $\forall r\in \mathcal{R}$, both the head type set $T_{r,head}$ and the tail type set $T_{r,tail}$ are required to be non-empty. For the cases there the type set of an entity $e$ is empty ($e$ can be either head entity $e_h$ or the tail entity $e_t$), i.e., $T_{e}=\emptyset$, we assign uniform similarity scores, i.e., $s(e, r')=1, \forall r'\in \mathcal{R}$. %Note that $\sum_{r'\in \mathcal{R}}p(r'|\mathcal{T}(e_h, e_t))=1$. 
Different from the existing work ~\cite{ma2017transt} that measures semantic similarity by treating each type equally, our type-based prior probability with hierarchy-based type weights encode not only the type information but also the underlying hierarchies among types.  

$|T_{r,head}|$ or $|T_{r,tail}|$ of relation $r$ can be
large containing long tail noisy types, which may weaken the relation prediction performance of the prior model. To remove noisy types, we introduce a threshold $\eta$. Type $t\in T_{r,*}$ will be kept if 
\begin{equation}
    w_{r,*}(t) \geq W_{r,*}^{min} + \eta\times(W_{r,*}^{max}-W_{r,*}^{min})
\end{equation}
%$w_{r,*}(t) \geq W_{r,*}^{min} + \eta\times(W_{r,*}^{max}-W_{r,*}^{min})$, 
where $W_{r,*}^{max}$ and $W_{r,*}^{min}$ are the maximum and the minimal weight of the type set $T_{r,*}$ respectively. If $w_{r,*}(t)$ is not sufficiently high(i.e., Eq.6 is not satisfied), type $t$ will be removed from $T_{r,*}$ without further consideration. %and $w_{r,*}(t)$ will be set to be zero. 
$*=\{\text{head}, \text{tail}\}$. Threshold $\eta$ varies from datasets to datasets and is chosen based on the performance on the validation set. %and varies from datasets to datasets (See Appendix A for detailed analysis of $\eta$).
\subsection{Embedding Based Models} Embedding based models represent
relations and entities in a continuous embedding space. We denote the embedding of the head and tail entities as $\textbf{e}_h$ and $\textbf{e}_t$ respectively. The embedding of relation $r$ is $\textbf{r}$. A score function $f_r(\textbf{e}_h, \textbf{e}_t)$ is usually defined as a measurement of the salience of a triple $(e_h, r, e_t)$. Embeddings are then learned by optimizing the score function based on existing instance-level triples. We consider three embedding based models:
\begin{itemize}
    \item \textbf{TransE}~\cite{bordes2013translating}:
    \begin{equation}
        f_r(\textbf{e}_h, \textbf{e}_t) = -||\textbf{e}_h\cdot\textbf{r} - \textbf{e}_t|| 
    \end{equation}
    where $\textbf{e}_h, \textbf{e}_r\in \mathbb{R}^d$ and $\textbf{r}\in \mathbb{R}^d$.
    \item \textbf{RotatE}~\cite{sun2019rotate}:
    \begin{equation}
        f_r(\textbf{e}_h, \textbf{e}_t) = -||\textbf{e}_h\circ\textbf{r} - \textbf{e}_t||
    \end{equation}
    where $\textbf{e}_h, \textbf{e}_r\in \mathbb{C}^d$ and $\textbf{r}\in \mathbb{C}^d$.
    \item \textbf{QuatE}~\cite{zhang2019quaternion}:
    \begin{equation}
        f_r(\textbf{e}_h, \textbf{e}_t) = -||\textbf{e}_h\otimes \frac{\textbf{r}}{|\textbf{r}|} - \textbf{e}_t||
    \end{equation}
    where $\textbf{e}_h, \textbf{e}_r\in \mathbb{H}^d$ and $\textbf{r}\in \mathbb{H}^d$.
\end{itemize}
%\begin{equation}
%    \begin{split}
%        &\text{TransE:}\ f_r(\textbf{e}_h, \textbf{e}_t) = -||\textbf{e}_h\cdot\textbf{r} - \textbf{e}_t||\\
%        &\text{RotatE:}\ \\
%        &\text{QuatE:}\ f_r(\textbf{e}_h, \textbf{e}_t) = -||Q_h\otimes W_r^{norm} - Q_r||\\
%    \end{split}
%\end{equation}
All of these three models concentrate on existing triples to learn embedding vectors without considering any type information. The learned embeddings thus only contain instance-level information observed from existing triples. For each triple $(e_h, r, e_t)$, we define the
likelihood of relation based on their corresponding embeddings as
\begin{equation}
    p(e_h, e_t|r) \triangleq  \exp(f_r(\textbf{e}_h, \textbf{e}_t)) %\frac{e^{f_r(\textbf{e}_h, \textbf{e}_t)}}{\sum_{r'\in \mathcal{R}}e^{f_r'(\textbf{e}_h, \textbf{e}_t)}}
\end{equation}
It is intuitive that the likelihood of relation will be small with a low score  $f_r(\textbf{e}_h, \textbf{e}_t)$.
\subsection{Type Information Integration} In the end, we propose to integrate the type information into the embedding based models at the decision-level. Given a triple $(e_h, r, e_t)$, we obtain the prior probability of relation $r$ based on the type information, i.e., $p(r|\mathcal{T}(e_h, e_t, \mathcal{R}))$. On the other hand, we obtain the likelihood of relation based on embeddings learned from existing triples, i.e., $p(e_h, e_t|r)$. Combining them together, we obtain the posterior probability of relation $r$ by following the Bayes rule, i.e., 
\begin{equation}
    p(r|e_h, e_t,\mathcal{T}(e_h, e_t, \mathcal{R})) \propto p(e_h, e_t|r)p(r|\mathcal{T}(e_h, e_t, \mathcal{R}))
\end{equation}
The posterior probability $p(r|e_h, e_t,\mathcal{T}(e_h, e_t, \mathcal{R}))$ thus contains both the type information and the instance-level information. Different from existing work, e.g.,\cite{ma2017transt, xie2016representationT}, that integrates the type information at the feature-level, we propose to integrate the type information at the decision-level and our proposed integration approach is independent of the embedding based models. %Hence, type information can be directly combined with existing embedding based models without any additional training.

\section{Experiments}
\label{sec:experiments}

To evaluate the performance of our  type-augmented relation prediction  (TaRP) approach, we first perform ablation studies on the prior model; and then evaluate the performance of the TaRP model. We demonstrate the effectiveness of the TaRP model by comparing it to three baseline embedding based models: TransE, RotatE, and QuatE. In addition, we show that by incorporating type information, TaRP is much more data efficient than existing methods. Furthermore, we demonstrate the generalization ability of type information. In the end, we compare our approach to state-of-the-art models that also apply ontological information.

\paragraph{Datasets} We consider three benchmark datasets for the relation prediction task: FB15K~\cite{bordes2013translating}, YAGO26K-906~\cite{hao2019universal} and  DB111K-74~\cite{hao2019universal}. FB15K is a popular benchmark dataset for the KG completion task, and its type information has been explored by most of the prior work, such as~\cite{ma2017transt, guo2015semantically, xie2016representationT}. YAGO26K-906 and DB111K-906 are two very recent datasets containing explicit ontological information, and have not been widely considered by related work. 

FB15K consists of triples extracted from the FreeBase knowledge graph~\cite{bollacker2008freebase}. The same type information is applied as introduced in~\cite{xie2016representationT} for FB15K\footnote{The hierarchical type $\textbf{\textit{/common/topic}}$ is removed as this is applied to every entity.}. 
Both YAGO26K-906 and DB111K-174 contain two types of KGs: instance KG and ontology KG, which are connected to each other through type links. The instance KGs of YAGO26K-906 and DB111K-174 consist of triples extracted from the YAGO knowledge graph~\cite{rebele2016yago} and the DBpedia knowledge graph~\cite{lehmann2015dbpedia} respectively; and are applied for the relation prediction task. Type links and ontology KGs are collected as type information (see Appendix B for the details of the pre-processing). %In particular, type $t$ is of one lever higher than type $t'$ in the type hierarchy if $t$ is connected to $t'$ through the relation $\texttt{is\_a}$.  type hierarchies are extracted from the ontology knowledge graph and we only consider the \texttt{isa} relation. 
Statistical information about the three datasets is shown in Table~\ref{tb_dataset_statistics}. 
\begin{table}[h] 
\begin{center}
\centering
\caption{Statistics of dataset.}
\label{tb_dataset_statistics}
\renewcommand{\arraystretch}{1.}
\tabcolsep0.1in
%\scalebox{0.8}{
\adjustbox{max width=\textwidth}{
\begin{tabular}{|c|c|c|c|c|c|c|}
\hline
\hline
Dataset & $\#$Rel. & $\#$Ent. & $\#$Types & $\#$Train & $\#$Valid & $\#$Test\\
\hline
FB15K& 1,345 &14,951 & 663&483,142 &50,000 &59,071  \\
YAGO26K-906& 34 &26,078 &226 &351,664 &-- &39,074  \\
DB111K-174& 298 &98,336 &242 &592,654 &-- &65,851  \\\hline
\end{tabular}}
\end{center}
\end{table}

On all three datasets, for each relation, the obtained head type set and tail type set are non-empty. For each entity from FB15K and DB111K-74, the type set is non-empty. On YAGO26K-906, only 8,948 entities have non-empty type sets. As a result, $4,149 (10.6\%)$ testing triples have type information for both head and tail entities; $30,839 (78.9\%)$ triples have type information for only head entity or only tail entity; and $4,086 (10.5\%)$ triples don't have type information for both head and tail entities. For the cases where the type set of entity $e$ is empty ($e$ can be either head entity $e_h$ or tail entity $e_t$), we assign uniform similarity scores, i.e., $s(e, r')=1,\forall r'\in \mathcal{R}$. 

%among all the testing data, only 4,149 triples($10.6\%$) have type information for both head entity and tail entity. 30,839 triples($78.9\%)$ have type information for either head entity or tail entity. And 4,086 triples($10.5\%$) don't have type information. For the testing triple with no type information, uniform distribution over relations is assigned. 

\vspace{+0.1cm}
\noindent \textbf{Evaluation protocol} For each triple $(e_h, r, e_{t})$ in the testing set, we replace the relation $r$ with every relation $r'\in \mathcal{R}$. We calculate the posterior probabilities $p(r'|e_h, e_t,\mathcal{T}(e_h, e_t, \mathcal{R}))$ of all replacement triples and rank these probabilities in descending order. We apply the filter setting~\cite{ma2017transt}. Two standard measures are considered as evaluation matrices: mean rank (MR) and Hits$@N$. %as considered in~\cite{sun2019rotate}. 
A higher Hits$@N$ and a lower MR mean better performance. In all the experiments, we report both Hits$@1$ and Hits$@10$.  

\vspace{+0.1cm}
\noindent \textbf{Experimental Settings}  TaRP has one hyper-parameter threshold $\eta$. On each dataset, we select the threshold $\eta$ from $\{0, 0.1, 0.2, 0.4, 0.6, 0.8, 0.9\}$ that achieves the best relation prediction performance (Hits$@1$) on the validation set (See Appendix A for detailed analysis of $\eta$). On FB15K and YAGO26K-906, $\eta=0.1$. On DB111K-174, $\eta=0$. We report the averaged size of type set over all the entities or relations as shown in Table~\ref{tb_type_statistics}\footnote{$|T_{r,head}|$ and $|T_{r,tail}|$ are calculated with optimal thresholds applied}. 
For baseline embedding based models, we directly reuse the best configurations provided by previous studies \cite{sun2019rotate, zhang2019quaternion}.%\footnote{https://github.com/DeepGraphLearning/KnowledgeGraphEmbedding} and \cite{zhang2019quaternion}.%\footnote{https://github.com/cheungdaven/QuatE}.
\begin{table}[h] 
\begin{center}
\centering
\caption{Averaged size of the type set.}
\label{tb_type_statistics}
\renewcommand{\arraystretch}{1.}
\tabcolsep0.1in
\scalebox{0.9}{
\begin{tabular}{|c|c|cc|}
\hline
\hline
\multirow{2}{*}{Dataset}&Entity&\multicolumn{2}{|c|}{Relation}\\
& $|T_{e}|$ & $|T_{r,head}|$ &$|T_{r,tail}|$\\
\hline
FB15K&12 & 20 & 19\\
YAGO26K-906&9 &6 &5  \\
DB111K-174& 2 &7 &12  \\\hline
\end{tabular}}
\end{center}
\end{table}

\subsection{Ablation Studies on the Prior Model}

We perform ablation studies to show the effectiveness of: 1) the  hierarchy-based type weights; 2) the type information.

\vspace{+0.2cm}
\noindent \textbf{Effectiveness of hierarchy-based type weights} To demonstrate the effectiveness of the proposed hierarchy-based type weights, we consider uniform weights for comparison, and calculate the prior probabilities
based on types with uniform weights. We compare the relation prediction performance of the prior model with hierarchy-based weights to the performance of the prior model with uniform weights. Results are shown in Table~\ref{tb_weights} where the prior model with hierarchy-based weights achieves much better performance than the prior model with uniform weights, particularly on FB15K and DB111K-174 datasets. These results empirically demonstrate the effectiveness of the hierarchy-based type weights.
\begin{table}[h] 
\begin{center}
\centering
\caption{Effectiveness of the hierarchy-based type weights.}
\label{tb_weights}
\adjustbox{max width=\textwidth}{
\begin{tabular}{|c|ccc|ccc|  }
\hline
\hline
\multirow{2}{*}{Type weights}  & \multicolumn{3}{|c|}{FB15K} &
\multicolumn{3}{|c|}{YAGO26K-906}\\
& MR & Hits@1 &Hits@10& MR & Hits@1 &Hits@10\\ \hline
Uniform &26&4.95&43.88&3.92 &71.85&88.75  \\
Hierarchy-based&\textbf{2.9}&\textbf{64.10} &\textbf{97.10}& \textbf{3.34}&\textbf{79.60} &\textbf{88.60}  \\\hline
\multirow{2}{*}{Type weights} &
    \multicolumn{3}{|c|}{DB111K-174} \\
& MR & Hits@1 &Hits@10\\ \cline{1-4}
Uniform &13.67&19.62 &63.61  \\
Hierarchy-based&\textbf{2.6}&\textbf{55.00}&\textbf{96.60} \\ \cline{1-4}
\end{tabular}}
\end{center}
\end{table}

\noindent \textbf{Effectiveness of type information} To
study the effectiveness of the type information, we evaluate the relation prediction performance of the prior model by considering the type information of 1) only head entity (H); 2) only tail entity (T); 3) both head and tail entities (H+T). For YAGO26K-906, only a subset of triples contain type information for both head and tail entities ($4,149$ triples), and hence we perform the evaluation on this subset for fair comparison. 
\begin{table}[h] 
\begin{center}
\centering
\caption{Effectiveness of the type information.}
\label{tb_ablation_prior}
\adjustbox{max width=\textwidth}{
\begin{tabular}{|c|ccc|ccc|  }
\hline
\hline
    \multirow{2}{*}{Type Info.}  & \multicolumn{3}{|c|}{FB15K} &
    \multicolumn{3}{|c|}{YAGO26K-906}\\
    & MR & Hits@1 &Hits@10& MR & Hits@1 &Hits@10\\ \hline
H &23.2&8.00&46.90& 4.2&56.20 &86.30  \\
T &20.3&9.00&50.20 &2.5&65.20 &95.70  \\\hline
H+T&\textbf{2.9}&\textbf{64.10} &\textbf{97.10}& \textbf{1.7}&\textbf{71.40} &\textbf{99.40}  \\\hline
\multirow{2}{*}{Type Info.} &
    \multicolumn{3}{|c|}{DB111K-174} \\
& MR & Hits@1 &Hits@10\\ \cline{1-4}
H &8.9&7.00&70.80  \\
T &12.5&19.70&62.10  \\\cline{1-4}
H+T&\textbf{2.6}&\textbf{55.00}&\textbf{96.60} \\ \cline{1-4}
\end{tabular}}
\end{center}
\end{table}
Results in Table~\ref{tb_ablation_prior} shows that considering the type information of head and tail entities jointly, the prior model achieves the best performance. These results depict that both the type information of head and tail entities are effective in relation prediction.%% and the prior model achieves the best performance when they are considered jointly. %In particular, on FB15K dataset, the Hits$@1$ of the prior model is very low if we apply the type information of only head entities or only tail entities. 

\subsection{Evaluation of the TaRP Model} 
\label{subsec:evaluation_of_tarp}

We evaluate the TaRP model by first comparing it to three baseline embedding based models. In addition, we show the data efficiency of the proposed TaRP model by reducing the number of training triples. More importantly, we perform cross-dataset evaluation and empirically demonstrate the generalization ability of the type information. %extracted from a certain dataset can generalized well to different datasets. 

\vspace{+0.2cm}
\noindent \textbf{Comparisons to baseline models}
As introduced in Section 3, we consider three baseline embedding based models: TransE, RotatE, and QuatE. The embeddings of entities and relations are obtained by directly running baseline models with reported best hyper-parameter settings. In addition, to demonstrate the effectiveness of the proposed decision-level integration, we enrich the existing training sets by adding type information as addition training triples; and train the embedding based models on enriched training sets for comparison. On YAGO26K-906 and DB111K-174, triples from ontology KG and type links can be directly used as additional training triples. On FB15K, given an entity $e$ and its hierarchical type $/t_1/t_2/.../t_K$, we collect type triples as $(e, r_1, t_K)$ with $r_1=\texttt{type}$ and $\{(t_k, r_2, t_{k-1})\}_{k=2}^K$ with $r_2=\texttt{is\_a}$. The embedding based models trained on enriched training sets can thus learn embeddings based on both existing triples and the type information. In other words, the type information is
fused with instance-level information at the feature-level. We denote the embedding based models learned from enriched training sets as: TransE(w/Type), RotatE(w/Type) and QuatE(w/Type). By combining the prior model with three embedding based models separately, we obtain three TaRP models: TaRP-T, TaRP-R, and TaRP-Q. The results are shown in Table~\ref{tb_evaluate_posterior} (see Appendix C for additional results). 

\begin{table*}[ht!]
\caption{Evaluation of the Type-augmented Relation Prediction(TaRP) model}
%\vspace{-0.2cm}
\label{tb_evaluate_posterior}
\centering
%\renewcommand{\arraystretch}{0.9}  
%\adjustbox{max width=\textwidth}{
\scalebox{0.9}{
\begin{tabular}{|c|ccc|ccc|ccc|}
\hline
\multirow{2}{*}{Method}   & \multicolumn{3}{|c|}{FB15K} & \multicolumn{3}{|c|}{YAGO26K-906} & \multicolumn{3}{|c|}{DB111K-174}\\ 
& MR & Hits@1 &Hits@10& MR  & Hits@1 &Hits@10& MR  & Hits@1 &Hits@10\\ \hline
%\multicolumn{2}{|c|}{Prior model} & 2.91 &64.10 &97.10 &3.27 &79.30&88.80 &2.64 & 55.00 &96.60\\ \hline
 TransE & 3.64 &76.50 &92.30 &1.12 &90.70&99.92 &4.76  & 66.60 & 86.70\\
RotatE & 2.38 & 80.20 & 97.80 & 1.10 & 92.84 & 99.90 & 4.53 & 65.90 & 93.80\\
QuatE & 4.01 & 82.20 & 94.90 & 1.33 & 91.65 & 98.96 & 8.56 & 58.60 & 88.90\\ \hline
TransE(w/Type) &3.32&79.37&91.56 &1.12&90.70&99.93 &4.16&67.64&91.91\\
RotatE(w/Type) &3.67 &73.63 &96.44 &\textbf{1.08}&\textbf{93.31}&99.93 &3.47 &70.08 &96.42\\
QuatE(w/Type) &3.98&80.82&92.97 &1.32&91.98&99.09&7.63&60.49&89.14\\\hline \hline
TaRP-T & 1.84 &88.90 &99.00 &1.10 &90.80&\textbf{99.98} &1.61  & 74.80 & 99.40\\
TaRP-R & \textbf{1.16} & \textbf{92.91} & \textbf{99.84} & \textbf{1.08} & 92.84 & \textbf{99.98} & \textbf{1.52} & 76.50 & \textbf{99.50}\\
TaRP-Q & 1.64 & 91.60 & 99.50 & 1.14 & 92.93 & 99.79 & 1.56 & \textbf{76.60} & 99.40\\ \hline 
\end{tabular}}%}
%\vspace{-0.15cm}
\end{table*}

From Table~\ref{tb_evaluate_posterior}, we can see that all three TaRP models achieve performance improvement on all three benchmark datasets compared to the corresponding baseline embedding based models. In particular, on FB15K and DB111K-174, the improvement is significant. For instance, TaRP-R obtains $92.91\%$ for Hits$@1$ on FB15K, achieving $12.71\%$ improvement compared to RotatE. On the other hand, on both YAGO26K-906 and DB111K-174, embedding based models trained on enriched training sets achieve improved performance compared to baseline embedding based models. However, for most of the embedding based models trained on enriched training sets, the achieved performance improvement is not as significant as the improvement achieved by the proposed TaRP model. TaRP models achieve overall better performance than the embedding based models trained on enriched training sets. For example, on DB111K-174, QuatE(w/Type) obtains $60.49\%$ for Hits$@1$; though higher than the $58.60\%$ obtained by QuatE, this is still significantly worse than the $76.60\%$ obtained by TaRP-Q. In addition, on FB15K, the embedding-based models trained with type triples perform worse than the embedding-based models trained without type triples. The reason may be that type triples collected from FB15K can contain errors\footnote{For example, for type \textbf{\textit{/book/author}}, the collected type triple (\textbf{\textit{author}}, \texttt{is\_a}, \textbf{\textit{book}}) is not true.} which leads to decreased performance. Our proposed prior model directly applies the type information, and hence errors introduced by type triples do not affect the TaRP models. 

These results show that by incorporating the type information, the TaRP model can always achieve better performance with different baseline embedding based models. More importantly, the TaRP models achieve overall better performance than the embedding based models trained with type triples, indicating that the proposed decision-level integration procedure is a more effective integration. In addition, through the proposed integration approach, the type information can be directly combined with embedding based models without additional training. 

\vspace{+0.2cm}
\noindent\textbf{Data efficiency of the TaRP model} 
%We evaluate the proposed TaRP model in terms of the data efficiency. In particular, 
We consider the embeddings that are learned from a small subset of training triples. Given insufficient training data, the quality of the learned embeddings will be lower. We compare the TaRP model where embeddings are learned from only a subset of training triples to the embedding based model that is trained on complete training sets. We perform this evaluation on FB15K and DB111K-174. RotatE is applied as the baseline embedding based model. We extract the subset of training triples with respect to each relation individually. %Four subsets of the training sets are considered:\{ $20\%$, $40\%$, $60\%$, $80\%$\}. The corresponding TaRP models are denoted as: TaRP-R($20\%$ D), TaRP-R($40\%$ D), TaRP-R($60\%$ D) and TaRP-R($80\%$ D), where D indicates the completed training data.   %We split the training triples w.r.t. the relations and collect a subset of training triples per each relation. 
Results are shown in Table~\ref{tb_ablation_reduceData}. 
\begin{table}[!ht]
\caption{Data efficiency of the TaRP model(D:training data)}
%\vspace{-0.2cm}
\label{tb_ablation_reduceData}
\centering
\adjustbox{max width=\textwidth}{
%\scalebox{0.8}{
    \begin{tabular}{|c|ccc|ccc|}
    \hline
    \multirow{2}{*}{Method}  & \multicolumn{3}{|c|}{FB15K} & \multicolumn{3}{|c|}{DB111K-174}\\ 
   & MR & Hits@1 &Hits@10& MR  & Hits@1 &Hits@10\\ \hline
   RotatE(100\% D) &2.38&80.20 & 97.80 & 4.53 & 65.90 & 93.80\\\hline
    TaRP-R(20\% D) &1.90 &83.20 &99.20 &2.17&63.40&97.60\\
    TaRP-R(40\% D) &1.74&85.90&99.60 &1.78 & 70.90&98.70\\
    TaRP-R(60\% D)&1.73&84.90&99.70&1.62&74.80 &99.10\\ 
    TaRP-R(80\% D) &1.71&85.50&99.70&1.55&76.00&99.40\\ \hline 
    \end{tabular}}%}
%\vspace{-0.15cm}
\end{table}
As shown, on FB15K, by integrating the type information, TaRP-R achieves better performance than RotatE with only $20\%$ of the training data. On DB111K-174,  TaRP-R achieves better performance with $40\%$ of the training data. These results show that TaRP achieves better performance than the embedding-based model using much lesser training data. By leveraging type information, TaRP has lesser dependency on the training triples, and thus is more data efficient.

\vspace{+0.2cm}
\noindent \textbf{Cross-dataset evaluation of the TaRP model} 
%\FloatBarrier
\begin{table*}[htb!]
\caption{Cross-dataset evaluation of the TaRP model}
%\vspace{-0.2cm}
\label{tb_evaluate_cross}
\centering
%\renewcommand{\arraystretch}{0.9}  
%\adjustbox{max width=\textwidth}{
\scalebox{0.9}{
\begin{tabular}{|c|ccc|ccc|ccc|}
\hline
\multirow{2}{*}{Method}   & \multicolumn{3}{|c|}{FB15K} & \multicolumn{3}{|c|}{YAGO26K-906} & \multicolumn{3}{|c|}{DB111K-174}\\ 
& MR & Hits@1 &Hits@10& MR  & Hits@1 &Hits@10& MR  & Hits@1 &Hits@10\\ \hline
RotatE &1.76&82.47&98.59 &1.09 & 92.55 & 99.92 &2.59 & 79.42 & 97.21 \\ \hline
TaRP-R(FB prior) &1.37 &92.88 &99.79 &1.06 &94.29&99.98 &2.06  &80.55 & 97.88\\
TaRP-R(YG prior) &1.74&85.77 &98.97 &1.05 &95.75 &99.99 & 2.22 & 79.90 & 97.62 \\
TaRP-R(DB prior) &5.99 &84.95  &98.25 &1.06 &94.62& 99.99 & 1.39 & 86.20 & 99.22\\ 
TaRP-R(Union prior)&1.19&92.90&99.83 &1.04&95.81&99.99 &1.34 &86.57 &99.40\\ \hline 
\end{tabular}}%}
%\vspace{-0.15cm}
\end{table*}
%\FloatBarrier
To demonstrate the generalization ability of the type information, we perform cross-dataset evaluation. In particular, the type information is collected from outside of the dataset. Given two knowledge graphs $\mathcal{G}^1=\{\mathcal{E}^1, \mathcal{R}^1, \mathcal{T}^1\}$ and $\mathcal{G}^2=\{\mathcal{E}^2, \mathcal{R}^2, \mathcal{T}^2\}$, for each entity $e\in \mathcal{E}^1$, we perform exact string matching to find if there exists a matched entity $e\in \mathcal{E}^2$. If so, we collect the type information for $e\in \mathcal{E}^1$ as $T_e = \{t_k\}_{k=1}^K$, where $t_k\in \mathcal{T}^2$ and $K$ is the total number of types for entity $e$.  %In the end, $3,162$ matched entities are found between YAGO26K-906 and FB15K, $9,393$ matched entities are found between DB111K-174 and FB15K, and $4,500$ matched entities are found between YAGO26K-906 and DB111K-174. 
For each dataset, we transfer the type information from the other two datasets individually. Given the transferred type information, we collect the relations whose head type set and tail type set are both non-empty in order to perform valid type information encoding. Only the testing triples that contain such relations are considered for the evaluation. 
In the end, on the FB15K dataset, $785$ qualified relations are obtained resulting in $55,804$ testing triples. On the YAGO26K-906 dataset, $32$ qualified relations are obtained resulting in $37,401$ testing triples. On the DB111K-27 dataset, $134$ qualified relations are obtained resulting in $55,037$ testing triples. For comparison, we extract the type information from within-dataset. The prior models with type information extracted from FB15K, YAGO26K-906, and DB111K-174 are denoted as FB prior, YG prior, and DB prior respectively. In addition, for each dataset, we consider a union prior where we combine the type sets of each entity extracted from the other two datasets with the existing type set collected from within-dataset. The embedding based models are directly trained on the training triples from within-dataset. RotatE is applied as the baseline embedding based model. For each dataset, we combine the baseline embedding based model with four different prior models individually, resulting in four TaRP models: TaRP-R(FB prior), TaRP-R(YG prior), TaRP-R(DB prior) and TaRP-R(Union prior). Results are shown in Table~\ref{tb_evaluate_cross}. As we can see from the Table~\ref{tb_evaluate_cross}, TaRP-R with type information collected from cross-datasets can still achieve performance improvement compared to the baseline embedding based model. For example, on FB15K, $3.3\%$ improvement is achieved with the type information transferred from YAGO26K-906. Furthermore, the TaRP-R with union-prior achieves better performance than the TaRP-R with type information collected within-datasets by leveraging additional type information collected from cross-datasets. From these results, we can see that the type information extracted from a specific dataset can generalize well to different datasets. In addition, through the proposed decision-level integration, the embedding based model can be easily combined with different type information without additional training.

\subsection{Comparisons to State-Of-The-Art Methods}

We compare TaRP to additional SoTA models that also apply ontological information. In particular, on FB15K, we compared to DKRL~\cite{xie2016representationD}, TKRL~\cite{xie2016representationT}, SSP~\cite{xiao2017ssp}, and TransT~\cite{ma2017transt}. TKRL and TransT apply type information. DKRL and SSP apply contextual information like descriptions of entities. The results are shown in Table~\ref{tb_SOTA_fb}. $^*$ indicates the reported performance.  On YAGO26K-906 and DB111K-174, we compare to the state-of-the-art model, JOIE~\cite{hao2019universal}. We train JOIE on two datasets with its reported best hyper-parameter configurations, and the results are shown in Table~\ref{tb_SOTA}. From Table~\ref{tb_SOTA_fb} and Table~\ref{tb_SOTA}, we can see that TaRP-R achieves the best performance, in particular for Hits$@10$. By integrating type information, the ranks of triples are concentrated within $\text{rank} 1$- $\text{rank} 10$. Hence, TaRP-R achieves very high Hits$@10$ and significantly outperforms SoTA methods on all three datasets.
\begin{table}[!htbp]
\caption{Comparisons with SOTA on FB15k}
%\vspace{-0.2cm}
\label{tb_SOTA_fb}
\centering
\renewcommand{\arraystretch}{0.9}  
\adjustbox{max width=\textwidth}{
%\scalebox{0.8}{
\begin{tabular}{|c|ccc|}
\hline
Methods  & MR & Hits@1 &Hits@10\\ \hline
%DisMult~\cite{yang2014embedding} & \\
%ComplEx~\cite{trouillon2016complex} & \\
DKRL(CNN)+TransE~\cite{xie2016representationD} &$2.03^*$ &-& $90.8^*$\\
TKRL(RHE)~\cite{xie2016representationT} & $1.73^*$ & $92.8^*$ &-\\
SSP(Std.)~\cite{xiao2017ssp} & $1.22^*$ & - & $89.2^*$ \\
SSP(Joint)~\cite{xiao2017ssp} & $1.47^*$ & - & $90.9^*$\\
TransT~\cite{ma2017transt} & $1.19^*$ & - & $94.1^*$\\
%JOIE~\cite{hao2019universal} &-  &-  &-  &1.47&90.1&97.1&2.22&71.8&89.6\\\hline 
\hline
\textbf{TaRP-R} &\textbf{1.16} & \textbf{92.9} & \textbf{99.8} \\ %& 1.08 & 92.8 &99.9 & 1.52 & 76.5 & 99.5\\
\hline
\end{tabular}}%}
%\vspace{-0.15cm}
\end{table}

\begin{table}[!htbp]
\caption{Comparisons with SOTA on YAGO26K-906 and DB111K-174}
%\vspace{-0.2cm}
\label{tb_SOTA}
\centering
\renewcommand{\arraystretch}{0.9}  
\adjustbox{max width=\textwidth}{
%\scalebox{0.8}{
\begin{tabular}{|c|ccc|ccc|}
\hline
\multirow{2}{*}{Method}  &\multicolumn{3}{|c|}{YAGO26K-906}& \multicolumn{3}{|c|}{DB111K-174}\\ 
& MR & Hits@1 &Hits@10& MR  & Hits@1 &Hits@10\\ \hline
JOIE~\cite{hao2019universal} &1.47&90.1&97.1&2.22&71.8&89.6\\\hline 
\textbf{TaRP-R} & \textbf{1.08} & \textbf{92.8} &\textbf{99.9} & \textbf{1.52} & \textbf{76.5} & \textbf{99.5}\\
\hline
\end{tabular}}%}
%\vspace{-0.15cm}
\end{table}

\section{Discussion}
Though majority of the related works that are aligned well with our proposed method performed evaluations on FB15K(as shown in Table~\ref{tb_SOTA_fb}), FB15K contains several shortcomings, such as data leakage problem. To address the potential concerns on the evaluation regarding to the problems within the FB15K, we consider the FB15K-237~\cite{dettmers2018convolutional}, which is an improved version of FB15K. We perform two evaluations on FB15K-237: 1) compare the proposed TaRP model to the baseline models; 2) compare to the SOTA method. We firstly evaluate the effectiveness the proposed approach on FB15K-237 by comparing to baseline models. 
\begin{table}[ht!]
\caption{Evaluation of the TaRP model on FB15K-237}
%\vspace{-0.2cm}
\label{tb_evaluate_posterior_237}
\centering
%\renewcommand{\arraystretch}{0.9}  
%\adjustbox{max width=\textwidth}{
\scalebox{0.9}{
\begin{tabular}{|c|ccc|}
\hline
& MR & Hits@1 &Hits@10\\ \hline
 TransE & 1.51 &93.18 &98.27\\
RotatE & 1.88&93.89&99.18\\
QuatE & 1.65&90.83&98.58\\ \hline
TaRP-T & \textbf{1.17}&\textbf{94.64}&99.76\\
TaRP-R & 1.19&94.25&\textbf{99.79} \\
TaRP-Q & 1.24&92.51&99.73\\ \hline 
\end{tabular}}%}
%\vspace{-0.15cm}
\end{table}
As we can see from Table~\ref{tb_evaluate_posterior_237}, all three TaRP models achieve performance improvement on FB15K-237 compared to the corresponding baseline embedding based models. Particularly, MR is reduced significantly from $1.88$ to $1.19$ by augmenting the RotatE with type information through the proposed framework. We then compare the TaRP model to the SOTA model: HAKE~\cite{zhang2020learning}. We train HAKE with its reported hyper-parameter settings. From Table~\ref{tb_SOTA_237}, we can see that the TaRP-R significantly outperforms the HAKE model. 
\begin{table}[!htbp]
\caption{Comparisons with SOTA on FB15K-237}
%\vspace{-0.2cm}
\label{tb_SOTA_237}
\centering
\renewcommand{\arraystretch}{0.9}  
\adjustbox{max width=\textwidth}{
%\scalebox{0.8}{
\begin{tabular}{|c|ccc|}
\hline
& MR & Hits@1 &Hits@10 \\ \hline
HAKE~\cite{zhang2020learning} &1.85&92.85&99.13\\\hline 
\textbf{TaRP-R} & \textbf{1.19} & \textbf{94.25} &\textbf{99.79} \\
\hline
\end{tabular}}%}
%\vspace{-0.15cm}
\end{table}

\section{Conclusion}
\label{sec:conclusion}

In this paper, we propose an effective type-augmented relation prediction (TaRP) method, where we apply both type information and instance-level information for relation prediction in knowledge graphs. The type information and instance-level information are encoded as prior probabilities and likelihoods of relations respectively, and are combined at the decision-level. Our approach significantly outperforms state-of-the-art methods. Additionally, by leveraging type information, the TaRP model is able to be more data efficient than existing models. Furthermore, type information extracted from a specific dataset is shown to generalize well to other datasets.

\section*{Acknowledgement}
This work is supported by the Rensselaer-IBM AI Research Collaboration (http://airc.rpi.edu), part of the IBM AI Horizons Network (http://ibm.biz/AIHorizons). Part of this work is also supported by DARPA grant FA8750-17-2-0132. 

\appendix

\section{Effectiveness of the threshold $\eta$}
In practice, $|T_{r,head}|$ or $|T_{r,tail}|$ can be a  large set containing long tail noisy types, which may weaken the performance of the prior model. To remove noisy types, we introduce a threshold $\eta$. Type $t\in T_{r,*}$ will be kept if 
\begin{equation}
    w_{r,*}(t) \geq W_{r,*}^{min} + \eta\times(W_{r,*}^{max}-W_{r,*}^{min})
\end{equation}
%$w_{r,*}(t) \geq W_{r,*}^{min} + \eta\times(W_{r,*}^{max}-W_{r,*}^{min})$, 
where $W_{r,*}^{max}$ and $W_{r,*}^{min}$ are the maximun and the minimal weight of the type set $T_{r,*}$ respectively. If $w_{r,*}(t)$ is not sufficiently high(i.e., Eq.1 is not satisfied), type $t$ will be removed from $T_{r,*}$ without further consideration. %and $w_{r,*}(t)$ will be set to be zero. 
$*=\{\text{head}, \text{tail}\}$. Threshold $\eta$ is chosen based on the relation prediction performance on the validation set and varies from datasets to datasets.  On each dataset, we manually select the optimal threshold $\eta$ from $\{0, 0.1, 0.2, 0.4, 0.6, 0.8, 0.9\}$ that achieves the best performance(Hits$@1$) on the validation set. On FB15K and YAGO26K-906, $\eta=0.1$. On DB111K-174, $\eta=0$. Relation prediction performance of the prior model with different thresholds $\eta$ on the validation set is shown in Table~\ref{tb_eta_fb}, Table~\ref{tb_eta_yago} and Table~\ref{tb_eta_DB} for FB15K, YAGO26K-906 and DB111K-174 respectively.

\begin{table}[ht] 
\begin{center}
\centering
\caption{Effectiveness of the threshold $\eta$ on FB15K}
\label{tb_eta_fb}
\adjustbox{max width=\textwidth}{
\begin{tabular}{|c|cc|ccc|}
\hline
\hline
Threshold &Ave.$|T_{r,head}|$ & Ave.$|T_{r,tail}|$ &MR & Hits$@1$ & Hits$@10$\\
\hline
0.0 & 46 & 43 & 3.47 & 52.2 & 95.6\\
0.1 & 20 & 19 & 2.78 & \textbf{64.2} & 96.6 \\
0.2 & 14 & 14 & \textbf{2.68} & 60.8 & \textbf{97.2}\\
0.4 & 7 & 7 & 3.56 & 57.8 & \textbf{97.2}\\
0.6 & 5 & 5 & 5.11 & 50.4 & 96.6 \\
0.8 & 4 & 4 & 6.12 & 46.8 & 97.0 \\
0.9 & 3 & 3 & 9.04 & 42.2 & 96.8\\
\hline
\end{tabular}}
\end{center}
\end{table}

%On both YAGO26K-906 and DB111K-174, there is no validation dataset and we consider the relation prediction performance of the prior model with different thresholds on the testing set, as shown in 
\begin{table}[ht] 
\begin{center}
\centering
\caption{Effectiveness of the threshold $\eta$ on YAGO26K-906}
\label{tb_eta_yago}
\adjustbox{max width=\textwidth}{
\begin{tabular}{|c|cc|ccc|}
\hline
\hline
Threshold &Ave.$|T_{r,head}|$ & Ave.$|T_{r,tail}|$ &MR & Hits$@1$ & Hits$@10$\\
\hline
0.0 & 55 & 59 & \textbf{3.27} & 79.3 & \textbf{89.2}\\
0.1 & 6 & 5 & 3.34 & \textbf{79.6} & 88.6 \\
0.2 & 4 & 3 & 5.24 & 43.2 & 81.1\\
0.4 & 2 & 1 & 6.92 & 40.3 & 80.6\\
0.6 & 1 & 1 & 8.86 & 15.1 & 80.7 \\
0.8 & 1 & 1 & 8.90 & 15.1 & 80.4\\
0.9 & 1 & 1 & 9.52 & 9.2 & 80.4\\
\hline
\end{tabular}}
\end{center}
\end{table}

\begin{table}[ht] 
\begin{center}
\centering
\caption{Effectiveness of the threshold $\eta$ on DB111K-174}
\label{tb_eta_DB}
\adjustbox{max width=\textwidth}{
\begin{tabular}{|c|cc|ccc|}
\hline
\hline
Threshold &Ave.$|T_{r,head}|$ & Ave.$|T_{r,tail}|$ &MR & Hits$@1$ & Hits$@10$\\
\hline
0.0 & 7 & 12 & \textbf{2.64} & \textbf{54.9} & \textbf{96.6}\\
0.1 & 2 & 3 & 13.76 & 51.5 & 89.1 \\
0.2 & 2 & 2 & 25.77 & 50.6 & 81.1\\
0.4 & 1 & 2 & 28.30 & 47.8 & 71.2\\
0.6 & 1 & 1 & 28.60 & 21.3 & 56.6\\
0.8 & 1 & 1 & 30.14 & 21.5 & 56.3\\
0.9 & 1 & 1 & 30.74 & 21.5 & 58.1\\
\hline
\end{tabular}}
\end{center}
\end{table}

\section{Type collection for YAGO26K-906 and DB111K-174}
Both YAGO26K-906 and DB111K-174 contains two KGs: instance KG $\mathcal{G}^{inst} = \{\mathcal{E}^{inst}, \mathcal{R}^{inst}\}$ and ontology KG $\mathcal{G}^{onto} = \{\mathcal{T}^{inst}, \mathcal{R}^{onto}\}$. Two KGs are connected to each other through type links. Type links and ontology KG are collected as prior type information. In particular, for each entity $e\in \mathcal{E}^{inst}$, if there is a type link connecting entity $e$ to type $t\in \mathcal{T}^{onto}$, we have the type $t$ as the most specific type regarding to entity $e$. Given type $t\in \mathcal{E}^{onto}$, if there exists a triple $(t, is\_a, t')\in \mathcal{G}^{onto}$, where $t'\in \mathcal{T}^{onto}$ and $is\_a \in \mathcal{R}^{onto}$. we take $t'$ as the parent of type $t$, which is of one level lower than type $t$ within the hierarchy. We then repeat the procedure finding the parent of type $t'$, and we stop when we reach the root type that doesn't have any parent in $\mathcal{G}^{onto}$. We denote the root type as $t^{root}$ and we collect the types with its hierarchy as $H=\{t^{root}/.../t'/t\}$ regarding to entity $e\in \mathcal{E}^{inst}$. 

\section{Evaluation of the TaRP model}
\subsection{Performance per relation} 
\begin{figure}[!ht]
    \centering
    \includegraphics[width=3in, height = 2in]{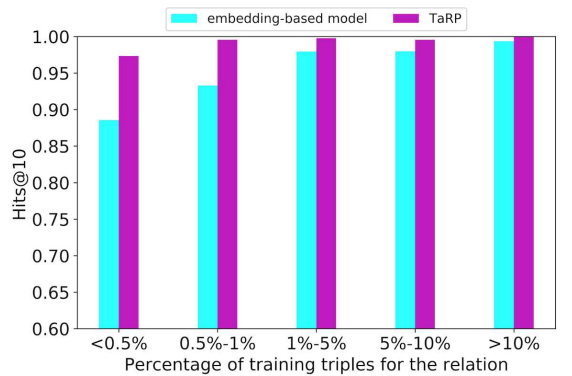}
    \caption{Performance per relation}
    \label{fig:perRelation}
\end{figure}

To further analyze the effectiveness of TaRP, we consider the performance of each relation individually. In particular, we categorize each relation $r$ based on the percentage of training triples containing the relation $r$. We consider five categories: $\{<0.5\%, 0.5\%-1\%, 1\%-5\%, 5\%-10\%, >10\%\}$. %and summarize the testing performance on all three datasets. 
For each category, we record the Hits$@10$ of each relation within the category from all three datasets, and report the averaged Hits$@10$. We take RotatE as the baseline embedding based model, and results are visualized in Figure~\ref{fig:perRelation}. We see that for the relations that are contained in very few training triples (e.g., $<0.5\%$, or $0.5\%-1\%$), the training data is not sufficient to learn robust embeddings. Thus the prediction performance is significantly worse than the performance of relations having sufficient training data (e.g., $5\%-10\%$ or $>10\%$). On the other hand, relations with insufficient training triples can benefit more from type information, and thus TaRP achieves more significant improvement for relations with insufficient data (e.g., $<0.5\%$). 

\subsection{Case Study} In addition, we report some specific examples to better illustrate the effectiveness of the proposed FaRP model. The examples are from FB15K. RotatE is applied as the embedding based model. The cases are listed in Table~\ref{tb_cases}.

\begin{table}[!ht] 
\begin{center}
\centering
\caption{Case Studies.}
\label{tb_cases}
\adjustbox{max width=\textwidth}{
\begin{tabular}{|ccc|}
\hline
\hline
\multicolumn{3}{|c|}{Rank}\\ \hline
 Prior model & Embedding-based model & TaRP\\
\hline
\multicolumn{3}{|c|}{(\texttt{Meet Dave}, \texttt{/people/person/gender}, \texttt{Male})}\\
 5 & 1169 & 10\\ \hline
\multicolumn{3}{|c|}{(\texttt{Police}, \texttt{/gardening\_hint/split\_to}, \texttt{Police procedural})}\\
 158 & 1333 & 469\\ \hline
\multicolumn{3}{|c|}{(\texttt{Jean-Claude Van Damme}, \texttt{/martial\_artist/martial\_art}, \texttt{Taekwondo})}\\
1 & 1345 & 6\\
\hline
\end{tabular}}
\end{center}
\end{table}

{\small
	\bibliography{egbib}
}
\end{document}